# Rule-based Machine Learning Methods for Functional Prediction

**Sholom M. Weiss**                                    WEISS@CS.RUTGERS.EDU
*Department of Computer Science, Rutgers University*
*New Brunswick, New Jersey 08903, USA*

**Nitin Indurkhya**                                    NITIN@CS.USYD.EDU.AU
*Department of Computer Science, University of Sydney*
*Sydney, NSW 2006, AUSTRALIA*

## Abstract

We describe a machine learning method for predicting the value of a real-valued function, given the values of multiple input variables. The method induces solutions from samples in the form of ordered disjunctive normal form (DNF) decision rules. A central objective of the method and representation is the induction of compact, easily interpretable solutions. This rule-based decision model can be extended to search efficiently for similar cases prior to approximating function values. Experimental results on real-world data demonstrate that the new techniques are competitive with existing machine learning and statistical methods and can sometimes yield superior regression performance.

## 1. Introduction

The problem of approximating the values of a continuous variable is described in the statistical literature as *regression*. Given samples of output (response) variable $y$ and input (predictor) variables $\mathbf{x} = \{x_1...x_n\}$, the regression task is to find a mapping $y = f(\mathbf{x})$. Relative to the space of possibilities, finite samples are far from complete, and a predefined model is needed to concisely map $\mathbf{x}$ to $y$. Accuracy of prediction, i.e. generalization to new cases, is of primary concern. Regression differs from classification in that the output variable $y$ in regression problems is *continuous*, whereas in classification $y$ is strictly categorical. From this perspective, classification can be thought of as a subcategory of regression. Some machine learning researchers have emphasized this connection by describing regression as "learning how to classify among continuous classes" (Quinlan, 1993).

The traditional approach to the problem is classical linear least-squares regression (Scheffe, 1959). Developed and refined over many years, linear regression has proven quite effective for many real-world applications. Clearly the elegant and computationally simple linear model has its limits, and more complex models may fit the data better. With the increasing computational power of computers and with larger volumes of data, interest has grown in pursuing alternative nonlinear regression methods. Nonlinear regression models have been explored by the statistics research community and many new effective methods have emerged (Efron, 1988), including projection pursuit (Friedman & Stuetzle, 1981) and MARS (Friedman, 1991). Methods for nonlinear regression have also been developed outside the mainstream statistics research community. A neural network trained by back-propagation (McClelland & Rumelhart, 1988) is one such model. Other models can be found in numerical analysis (Girosi & Poggio, 1990). An overview of many different regression models, with application to classification, is available in the literature (Ripley, 1993). Most of these methods produce solutions in terms of weighted models.

In the real-world, classification problems are more commonly encountered than regression problems. This accounts for the greater attention paid to classification than to regression. But many important problems in the real world are of the regression type. For instance, problems involving time-series usually involve prediction of real values. Besides the fact that regression problems are important on their own, another reason for the need to focus on regression is that regression methods can be used to solve classification problems. For example, neural networks are often applied to classification problems.

The issue of interpretable solutions has been an important consideration leading to development of "symbolic learning methods." A popular format for interpretable solutions is the disjunctive normal form (DNF) model (Weiss & Indurkhya, 1993a). Decision trees and rules are examples of DNF models. Decision rules are similar in characteristics to decision trees, but they also have some potential advantages: (a) a stronger model (b) often better explanatory capabilities. Unlike trees, DNF rules need not be mutually exclusive. Thus, their solution space includes all tree solutions. These rules are potentially more compact and predictive than trees. Decision rules may also offer greater explanatory capabilities than trees because as a tree grows in size, its interpretability diminishes.

Among symbolic learning methods, decision tree induction, using recursive partitioning, is highly developed. Many of these methods developed within the machine learning community, such as ID3 decision tree induction (Quinlan, 1986), have been applied exclusively to classification tasks. Less widely known is that decision trees are also effective in regression. The CART program, developed in the statistical research community, induces both classification and regression trees (Breiman, Friedman, Olshen, & Stone, 1984). These regression trees are strictly binary trees, a representation which naturally follows from intensive modeling using continuous variables.[1]

In terms of performance, regression trees often are competitive in performance to other regression methods (Breiman et al., 1984). Regression trees are noted to be particularly strong when there are many higher order dependencies among the input variables (Friedman, 1991). The advantages of the regression tree model are similar to the advantages enjoyed by classification trees over other models. Two principal advantages can be cited: (a) dynamic feature selection and (b) explanatory capabilities. Tree induction methods are extremely effective in finding the key attributes in high dimensional applications. In most applications, these key features are only a small subset of the original feature set. Another characteristic of decision trees that is often cited is its capability for explanation in terms acceptable to people. On the negative side, decision trees cannot represent compactly many simple functions, for example linear functions. A second weakness is that the regression tree model is discrete, yet predicts a continuous variable. For function approximation, the expectation is a smooth continuous function, but a decision tree provides discrete regions that are discontinuous at the boundaries. All in all though, regression trees often produce strong results, and for many applications their advantages strongly outweigh their potential disadvantages.

---

1. A comparative study (Fayyad & Irani, 1992) suggests that binary classification trees are somewhat more predictive even for categorical variables.

In this paper we describe a new method for inducing regression rules. The method takes advantage of the close relationship between classification and regression and provides a uniform and general model for dealing with both problems. Additional gains can be obtained by extending this method in a manner that preserves the strengths of the partitioning schemes while compensating for their weaknesses. Rules can be used to search for the most relevant cases, and a subset of these cases can help determine the function value. Thus, some of the model's interpretability can be traded off for better performance. Empirical results suggest that these methods are effective and can induce solutions that are often superior to decision trees.

## 2. Measuring Performance

The objective of regression is to minimize the distance between the sample output values, $y_i$ and the predicted values $y_i^{'}$. Two measures of distance are commonly used. The classical regression measure is equation 1, the average squared distance between $y_i$ and $y_i^{'}$, i.e. the variance. It leads to an elegant formulation for the linear least squares model. The mean absolute distance (deviation) of equation 2 is used in least absolute deviation regression, and is perhaps the more intuitive measure.

The mean absolute distance (deviation) of equation 2 is used in our studies. This is a measure of the average error of prediction for each $y_i$ over $n$ cases.

$$Variance = \frac{1}{n}\sum_{i=1}^{n}(y_i - y_i^{'})^2 \qquad (1)$$

$$MAD = \frac{1}{n}\sum_{i=1}^{n}|y_i - y_i^{'}| \qquad (2)$$

The regression problem is sometimes described as a signal and noise problem. The model is extended to include a stochastic component $\epsilon$ in equation 3. Thus, the true function may not produce a zero error distance. In contrast to classification where the labels are assumed correct, for regression the predicted $y$ values could be explained by a number of factors including a random noise component, $\epsilon$, in the signal, $y$.

$$y = f(x_1 \ldots x_n) + \epsilon \qquad (3)$$

Because prediction is the primary concern, estimates based on training cases alone are inadequate. The principles of predicting performance on new cases are analogous to classification, but here the mean absolute distance is used as the error rate. The best estimate of true performance of a model is the error rate on a large set of independent test cases. When large samples of data are unavailable, the process of train and test is simulated by random resampling. In most of our experiments, we used (10-fold) cross-validation to estimate predictive performance.

## 3. Regression by Tree Induction

In this section, we contrast regression tree induction with classification tree induction. Like classification trees, regression trees are induced by recursive partitioning. The solution takes the form of equation 4, where $R_i$ are disjoint regions, $k_i$ are constant values, and $y_j^i$ refers to the y-values of the training cases that fall within the region $R_i$.

$$if\ \mathbf{x} \subseteq R_i\ then\ f(\mathbf{x}) = k_i = median\{y_j^i\} \tag{4}$$

Regression trees have the same representation as classification trees except for the terminal nodes. The decision at a terminal node is to assign a case a constant $y$ value. The single best constant value is the median of the training cases falling into that terminal node because for a partition, the median is the minimizer of mean absolute distance. Figure 1 is an example of a binary regression tree. All cases reaching shaded terminal node 1 (x1≤3) are assigned a constant value of y=10.

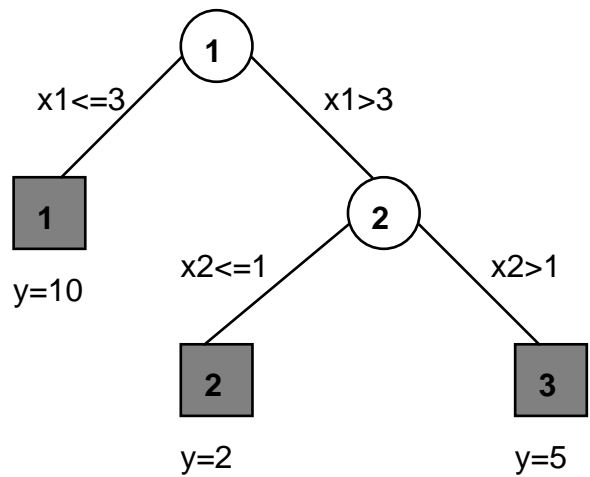

Figure 1: Example of Regression Tree

Tree induction methods usually proceed by (a) finding a covering set for the training cases and (b) pruning the tree to the best size. Although classification trees have been more widely studied, a similar approach can be applied to regression trees. We assume the reader is familiar with classification trees, and we cite only the differences in binary tree induction (Breiman et al., 1984; Quinlan, 1986; Weiss & Kulikowski, 1991). In many respects, regression tree induction is more straightforward. For classification trees, the error rate is a poor choice for node splitting, and alternative functions such as *entropy* or *gini* are employed. For regression tree induction, the minimized function, i.e. absolute distance, is most satisfactory. At each node, the single best split that minimizes the mean absolute distance is selected. Splitting continues until fewer than a minimum number of cases are covered by a node, or until all cases within the node have the identical value of y.

The goal is to find the tree that generalizes best to new cases, and this is often not a full covering tree, particularly in presence of noise or weak features. The pruning strategies employed for classification trees are equally valid for regression trees. Like the covering procedures, the only substantial difference is that the error rate is measured in terms of mean absolute distance. One popular method is the weakest-link pruning strategy (Breiman et al., 1984). For weakest-link pruning, a tree is recursively pruned so that the ratio *delta/n* is minimized, where *n* is the number of pruned nodes and *delta* is the increase in error.

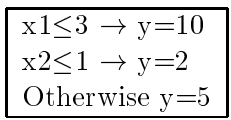

| |
|---|
| x1≤3 → y=10 |
| x2≤1 → y=2 |
| Otherwise y=5 |

Figure 2: Example of Regression Rules

Weakest link pruning has several desirable characteristics: (a) it prunes by training cases only, so that the remaining test cases are relatively independent (b) it is compatible with resampling.

## 4. Regression by Rule Induction

Both tree and rule induction models find solutions in disjunctive normal form, and the model of equation 4 is applicable to both. Each rule in a rule-set represents a single partition or region $R_i$. However, unlike the tree regions, the regions for rules need not be disjoint. With non-disjoint regions, several rules may be satisfied for a single sample. Some mechanism is needed to resolve the conflicts in $k_i$, the constant values assigned, when multiple rules, $R_i$ regions, are invoked. One standard model (Weiss & Indurkhya, 1993a) is to order the rules. Such ordered rule-sets have also been referred to as *decision lists*. The first rule that is satisfied is selected, as in equation 5.

$$if \ i < j \ and \ \mathbf{x} \subseteq both \ R_i \ and \ R_j \ then \ f(\mathbf{x}) = k_i \qquad (5)$$

Figure 2 is an example of an ordered rule-set corresponding to the tree of Figure 1. All cases satisfying rule 3, and not rules 1 and 2, are assigned a value of y=5.

Given this model of regression rule sets, the problem is to find procedures that effectively induce solutions. For rule-based regression, a covering strategy analogous to the classification tree strategy could be specified. A rule could be induced by adding a single component at a time, where each added component is the single best minimizer of distance. As usual, the constant value $k_i$ is the median of the region formed by the current rule. As the rule is extended, fewer cases are covered. When fewer than a minimal number of cases are covered, rule extension terminates. The covered cases are removed and rule induction can continue on the remaining cases. This is also the regression analogue of rule induction procedures for classification (Michalski, Mozetic, Hong, & Lavrac, 1986; Clark & Niblett, 1989).

However, instead of this approach, we propose a novel strategy of mapping the regression covering problem into a classification problem.

### 4.1 A Reformulation of the Regression Problem

The motivation for mapping regression into classification is based on a number of factors related to the extra information given in the regression problem: the natural ordering of $y_i$ by magnitude: if $i > j$ then $y_i > y_j$.

Let $\{C_i\}$ be a set consisting of an arbitrary number of classes, each class containing approximately equal values of $\{y_i\}$. To solve a classification problem, we expect that the classes are different from each other, and that patterns can be found to distinguish these

1. Generate a set of Pseudo-classes using the P-class algorithm (Figure 4).
2. Generate a covering rule-set for the transformed classification problem using a rule induction method such as Swap-1 (Weiss & Indurkhya, 1993a).
3. Initialize the current rule set to be the covering rule set and save it.
4. If the current rule set can be pruned, iteratively do the following:
   a) Prune the current rule set.
   b) Optimize the pruned rule set (Figure 5) and save it.
   c) Make this pruned rule set the new current rule set.
5. Use test cases or cross-validation to pick the best of the saved rule sets.

Figure 3: Overview of Method for Learning Regression Rules

classes. Should we expect classes formed by an ordering of $\{y_i\}$ to be a reasonable classification problem? There are a numbers of reasons why the answer is yes, particularly for a rule induction procedure.

The most obvious situation is the classical linear relationship. In this instance, by definition, some ordering of $\{x_{1i} \ldots x_{ni}\}$ corresponds to the ordering of $y_i$. Although classical methods are very strong in compactly determining linear functions, most interest in modern methods centers around their potential for finding nonlinear relationships. For nonlinear functions, we know there is usually no such ordering of $\{x_{1i} \ldots x_{ni}\}$ corresponding to the $\{y_i\}$. Still, we expect that the true function is smooth, and in a local region the ordering relationship will hold. In terms of classification, we know that a class $C_j$ with similar values of y is quite different than class $C_k$ with much lower values of y. For a nonlinear function within a class of similar values of y, some of these y have very similar values of $\{x_{1i} \ldots x_{ni}\}$. These correspond to some local region of the function. However, it is also true that some identical values of y can have very different $\{x_{1i} \ldots x_{ni}\}$ so that multiple clusters can be found within the class. Because rule induction methods do not cover a class with a single rule, the expectation is that multiple patterns will be found to cover these clusters.

Once the cases have been assigned such (pseudo-)classes, the classification problem can be solved in the following stages: (a) find a covering set and (b) prune the rule set to an appropriate size, with improved results achieved when an additional technique is considered: (c) refine or optimize a rule set. The overall method is outlined in Figure 3.

## 4.2 Generating Pseudo-classes

In the previous section, we described the motivation for pseudo-classes. The specification of these classes does not use any information beyond the ordering of $y$. No assumptions about the true nature of the underlying function are made. Within this environment, the goal is to make the $y$ values within one class most similar and $y$ values across classes most dissimilar. We wish to assign the $y$ values to classes such that the overall distance between each $y_i$ and its class mean is minimum.

Input: $\{y_i\}$ a set of output values
Initialize n := number of cases, k := number of classes

For each $Class_i$
    $Class_i$ := next n/k cases from list of sorted $y$ values
end-for

Compute $Err_{new}$
Repeat
    $Err_{old} = Err_{new}$
    For each $Case_j$
        When it is in $Class_i$
        1. If Dist$[Case_j,$ Mean$(Class_{i-1})]$ < Dist$[Case_j,$ Mean$(Class_i)]$
            Move $Case_j$ to $Class_{i-1}$
        2. If Dist$[Case_j,$ Mean$(Class_{i+1})]$ < Dist$[Case_j,$ Mean$(Class_i)]$
            Move $Case_j$ to $Class_{i+1}$
    Next $Case_j$
    Compute $Err_{new}$
Until $Err_{new}$ is not less than $Err_{old}$

Figure 4: Composing Pseudo-Classes (P-Class)

Figure 4 describes an algorithm (P-Class) for assigning the values $\{y_i\}$ to k classes. Essentially the algorithm does the following: (a) sorts the y values; (b) assigns approximately equal numbers of contiguous sorted $y_i$ to each class; (c) moves a $y_i$ to a contiguous class when that reduces the global distance Err from each $y_i$ to the mean of its assigned class. Classes with identical means should be merged. P-Class is a variation of k-means clustering, a statistical method that minimizes a distance measure (Hartigan & Wong, 1979). Alternative methods that do not depend on distance measures (Lebowitz, 1985) may also be used.

Given a fixed number of $k$ classes, this procedure will relatively quickly assign the $y_i$ to classes such that the overall distances are minimized. Because the underlying function is unknown, it is not critical to have a global minimum assignment of the $y_i$. This procedure matches well to our stated goals for ordering the $y_i$ values. The obvious remaining question is how do we determine $k$, the number of classes? Unfortunately, there is no direct answer, and some experimentation is necessary. However, as we shall see in Section 7, there is empirical evidence suggesting that results are quite similar within a local neighborhood of values of $k$. Moreover, relatively large values of $k$, which entail increased computational complexity for rule induction, are typically necessary only for noise-free functions that can be modeled exactly. Analogous to comparisons of neural nets with increasing numbers of hidden units, the trends for increasing numbers of partitions become evident during experimentation.

One additional variation on the classification theme arises for rule induction schemes that cover one class at a time. The classes must be ordered, and the last class typically becomes a default class to cover situations when no rule for other classes is satisfied. For regression, having one default partition for a class is unlikely to be the best covering solution, and instead the remaining cases for the last class are repeatedly partitioned (by P-Class) into 2 classes until fewer than $m$ cases remain.

An interesting characteristic of this transformation of the regression problem is that we now have a uniform and general model that once again relates both classification and regression. If the $y_i$ values are discrete and categorical, P-Class merely restates the standard classification problem. For example, if all values of $y_i$ are either 0 or 1, then the result of P-Class will be be 2 non-empty classes.

## 4.3 A Covering Rule Set

With this transformation, rule induction algorithms for classification can be applied. We will consider those induction methods that fully cover a class before moving on to induce rules for the next class. At each step of the covering algorithm, the problem is considered a binary classification problem for the current class $C_i$ versus all $C_j$ where $j > i$, i.e. the current class versus the remaining classes. When a rule is induced, its corresponding cases are removed and the remaining cases are considered. When a class has been covered, the next class is considered. An example of such a covering algorithm is that used in Swap-1 (Weiss & Indurkhya, 1993a), and this is the procedure used in this paper. The covering method is identical for classification and regression. However, one distinction is that the regression classes are transient labels that are replaced with the median of the y values for the cases covered by each induced rule. Because the rules are ordered and multiple rules may be satisfied, the medians are derived only from those instances where the rule is the first to be satisfied.

Although this procedure may yield good, compact covering sets, additional procedures are necessary for a complete solution.

## 4.4 Pruning the Rule Set

Typical real-world applications have noisy features that are not fully predictive. A covering set, particularly one composed of many continuous variables, can be far too over-specialized to produce the best results. For classification, relatively few classes are specified in advance. For regression, we expect many smaller groups because values of $y_i$ are likely to be quite different.

We noted earlier that for regression trees the usual classification pruning techniques can be applied with the substitution of mean absolute distance for the classification error rate. As in weakest-link tree pruning, the same ratio of $delta/n$ can be recursively minimized for weakest-link rule pruning. The intuitive rationale is to remove those parts of a rule set that have the least impact on increasing the error. Pruning rule sets is usually accomplished by either deleting complete rules or single rule components (Quinlan, 1987; Weiss & Indurkhya, 1993a). In general, rule pruning (for both classification and regression) is less natural and far more computationally expensive than tree pruning. Tree pruning has a natural flow from set to subset. Thus a tree can be pruned from bottom up, typically considering the effect of removing a subtree. Non-disjoint rules have no such natural pruning order, for example every component in a rule is a candidate for pruning and may affect all other rules that follow it in the specified rule order.

There is a major difference in pruning regression rules *vs.* classification rules. For classification, deleting a rule or a rule component has no effect on the class labels. For regression, pruning will change the median-values of y for the regions. Even the deletion of a rule will affect other region medians because the rules are ordered and multiple rules may be satisfied. This characteristic of rule pruning for regression adds substantial complexity to the task. However, by assuming that the median-values of y remain unchanged during the evaluation of candidate rules to prune, a pruning procedure can achieve reasonable computational efficiency at the expense of some loss in the accuracy of evaluation. Once the best rule or component for deletion is selected, the medians of all regions can then be re-evaluated.

Even for classification rules, rule pruning has some inherent weaknesses. For example, rule deletion will often create a gap in coverage. For classification rules though, it is quite feasible to develop an additional procedure to refine and optimize a rule set. To a large extent, this overcomes the cited weakness in pruned rules sets. A similar refinement and optimization procedure can be developed for regression and is described next.

## 4.5 Rule Refinement and Optimization

Given a rule set $RS_i$, can it be improved? This question applies to any rule set, although we are mostly motivated by trying to improve the pruned rules sets $\{RS_o \ldots RS_i \ldots RS_n\}$. This is a combinatorial optimization problem. Using error measure Err(RS), can we improve $RS_i$ without changing its size, i.e. the number of rules and components? Figure 5 describes an algorithm that minimizes Err(RS), the MAD of the model prediction on sample cases, by local swapping, i.e. replacing a single rule component with the best alternative. It is a variation of the techniques used in Swap-1 (Weiss & Indurkhya, 1993a).

The central theme is to hold a model configuration constant and make a single local improvement to that configuration. Local modifications are made until no further improvements are possible. Making local changes to a configuration is a widely-used optimization technique to approximate a global optimum and has been applied quite successfully, for example to find near-optimum solutions to traveling salesman problems (Lin & Kernighan, 1973). An analogous local optimization technique, called *backfitting*, has been used in the context of nonlinear statistical regression (Hastie & Tibshirani, 1990).

Variations on the selection of the next improvement move could include:

1. First local improvement encountered (such as in backfitting)

2. Best local improvement (such as in Swap-1)

In our experiments with rule induction methods, the results are consistently better for (2); (1) is more efficient, but the (pruned) rule induction environment is mostly stable with relatively few local improvements prior to convergence. In a less stable environment, with very large numbers of possible configuration changes, (2) may not be feasible or even better. In the pruned rule set environment, if the covering procedure is effective, then each pruned solution should be relatively close to a local minimum solution. Weakest-link pruning

Input: RS a rule set consisting of rules $R_i$, and
        S a set of training cases

D := TRUE
while (D is TRUE) do
        $RS_{new}$ := RS with the single best replacement for a
                component of RS that most reduces Err(RS) on
                cases in S using current Median($R_i$)
        If no replacement is found then
            D := FALSE
        else
            RS := $RS_{new}$; recompute Medians($R_i$)
endwhile
return the rule set RS

Figure 5: Optimization by Rule Component Swapping

results in a series of pruned rule sets $RS_i$ that number far fewer than sets which would result from a single prune of a rule or rule component. Each of the $RS_i$ are optimized prior to continuing the pruning process. However, rule set optimization can usually be suspended until substantial segments of the covering set have already been pruned.

If (1) is used, then either sequentially ordered evaluations (as in backfitting) or stochastic evaluations can be considered. Empirical evidence in the optimization literature supports the superiority of stochastic evaluation (Jacoby, Kowalik, & Pizzo, 1972). Further improvements may be obtained by occasionally making random changes in configuration (Kirpatrick, Gelatt, & Vecchi, 1983). These are general combinatorial optimization techniques that must be substantially reworked to fit a specific problem type. Most are expected to be applied throughout problem solving.

The result of pruning a covering rule set, $RS_o$, is a series of progressively smaller rule sets $\{RS_o \ldots RS_i \ldots RS_n\}$. The objective is to pick the best one, usually by some form of error estimation. Model complexity and future performance are highly related. Both too complex or too simple a model can yield poor results, the objective being to find just the right size model. Independent test cases or resampling by cross-validation are effective for estimating future performance. In the absence of these estimates, approximations, such as GCV (Craven & Wahba, 1979; Friedman, 1991), as described in equation 6, have been used in the statistics literature to estimate performance[2]. Both measures of training error and model complexity are used in the estimates. C(M), is a measure of model complexity expressed in terms of parameters estimated (such as the number of weights in a neural net) or tests performed, where C(M) is assumed to be less than n, the number of cases.

---

2. GCV is an acronym for generalized cross-validation, but only the apparent error on training cases is used and not true cross-validation by resampling.

$$GCV(M) = \sum_{i=1}^{n} \frac{\frac{|y_i - y_i'|}{n}}{1 - \frac{C(M)}{n}} \qquad (6)$$

In our experiments we used cross-validated estimates to guide the final model selection process, but other measures such as GCV may also be used.

## 4.6 Potential Problems with Rule-based Regression

Regression rules, like trees, are induced by recursive partitioning methods that approximate a function with constant-value regions. They are relatively strong in dynamic feature selection in high-dimensional applications, sometimes using only a few highly predictive features. An essential weakness of these methods is the approximation of a partition or region by a constant value. For a continuous function and even a moderately sized sample, this approximation can lead to increased error.

To deal with this limitation, instead of constant-value functions, linear functions can be substituted in a partition (Quinlan, 1993). However, a linear function has the obvious weakness that the true function may be far from linear even in the restricted context of a single region. In general, use of such linearity compromises the highly non-parametric nature of the DNF model. A better strategy might be to examine alternative non-linear methods.

## 5. An Alternative to Rules: k-Nearest Neighbors

The k-nearest neighbor method is one of the simplest regression methods, relying on table lookup. To classify an unknown case $\mathbf{x}$, the $k$ cases that are closest to the new case are found in a sample data base of stored cases. The predicted $y(\mathbf{x})$ of equation 7 is the mean of the $y$ values for the k-nearest neighbors. The nearest neighbors are found by a distance metric such as euclidean distance (usually with some feature normalization). The method is non-parametric and highly non-linear in nature

$$y_{knn}(\mathbf{x}) = \frac{1}{K} \sum_{k=1}^{K} y_k \;\; for \; K \; nearest \; neighbours \; of \; \mathbf{x} \qquad (7)$$

A major problem with this approach is how to limit the effect of irrelevant features. While limited forms of feature selection are sometimes employed in a preprocessing stage, the method itself cannot determine which features should be weighted more than others. As a result, the procedure is very sensitive to the distance measure used. In a high-dimensional feature space, k-nearest neighbor methods may perform very poorly. These limitations are precisely those that the partitioning methods address. Thus, in theory, the two methods potentially complement one another.

## 6. Model Combination

In practice, one learning model is not always superior to others, and a learning strategy that examines the results of different models may do better. Moreover, by combining different models, enhanced results may be achieved. A general approach to combining learning models is a scheme referred to as *stacking* (Wolpert, 1992). Additional studies have been performed in applying the scheme to regression problems (Breiman, 1993; LeBlanc & Tibshirani, 1993). Using small training samples of simulated data, and linear combinations of regression methods, improved results were reported. Let $M_i$ be the i-th model trained on the same sample, and $w_i$, the weight to be given to $M_i$.[3] If the new case vector is $\mathbf{x}$, the predictions of different models can be combined as in Equation 8 to produce an estimate of $y$. The models may use the same representation, such as k-nearest neighbors with variable-size $k$, or perhaps variable-size decision trees. The models could also be completely different, such as combining decision trees with linear regression models. Different models are applied independently to find solutions, and later a weighted vote is taken to reach a combined solution. This method of model combination is in contrast to the usual approach to evaluation of different models, where the single best performing model is selected.

$$y = \sum_{k=1}^{K} w_k M_k(\mathbf{x}) \qquad (8)$$

While stacking has been shown to give improved results on simulated data, a major drawback is that properties of the combined models are not retained. Thus when interpretable models are combined, the result may not be interpretable at all. It is also not possible to compensate for weaknesses in one model by introducing another model in a controlled fashion.

As suggested earlier, partitioning regression methods and k-nearest neighbor regression methods are complementary. Hence one might expect that by suitably combining the two methods, one might obtain better performance. In one recent study (Quinlan, 1993), model trees (i.e., regression trees with linear combinations at the leaf nodes) and nearest neighbor methods were also combined. The combination method is described in equation 9, where the $N(\mathbf{x})^k$ is one of the $K$ nearest neighbors of $\mathbf{x}$, $V(\mathbf{x})$ is the y-value of the stored instance $\mathbf{x}$, and $T(\mathbf{x})$ is the result of applying a model tree to $\mathbf{x}$.

$$y = \frac{1}{K} \sum_{k=1}^{K} V(N(\mathbf{x})^k) - (T(N(\mathbf{x})^k) - T(\mathbf{x})) \qquad (9)$$

The k-nearest neighbors are found independently of the induced regression tree (results were reported with $K=3$). In that sense, the approach is similar to the combination method of equation 8. The k-nearest neighbors are passed down the tree, and the results are used to refine the nearest neighbor answer. Thus, we have a combination model formed by independently computing a global solution, and later combining results.

However, there are strong reasons for not determining the global nearest neighbor solution independently. While, at the limit, with large samples, the non-parametric k-nearest neighbor methods will correctly fit the function, in practice though, their weaknesses can be substantial. Finding an effective global distance measure may not be easy, particularly in the presence of many noisy features. Hence a different technique for combining the two methods is needed.

---

3. These weights are obtained so as to minimize the least squared error under some constraints (Breiman, 1993).

## 6.1 Integrating Rules with Table-lookup

Consider the following strategy: To determine y-value of a case $\mathbf{x}$ that falls in region $R_i$, instead of assigning a single constant value $k_i$ for region $R_i$, where $k_i$ is determined by the median $y$ value of training cases in the region, assign $y_{knn}^i(\mathbf{x})$, the mean of the k-nearest (training set) instances of $\mathbf{x}$ in region $R_i$. Thus for regression trees, we now have equation 10. For regression rules, we also have equation 11.

$$if\ \mathbf{x} \subseteq R_i\ then\ f(\mathbf{x}) = y_{knn}^i(\mathbf{x}) \tag{10}$$

$$if\ i < j\ and\ \mathbf{x} \subseteq both\ R_i\ and\ R_j\ then\ f(\mathbf{x}) = y_{knn}^i(\mathbf{x}) \tag{11}$$

An interesting aspect of this strategy is that k-nearest neighbor results need only be considered for the cases covered by a particular partition. While this increases the interaction between the models and eliminates the independent computation of the two models, the model rationale and, as we shall show, the empirical results, are supportive of this approach.

We now have a representation which potentially alleviates the weakness of partitions being assigned single constant values. Moreover, some of the global distance measure difficulties of the k-nn methods may also be relieved because the table lookup is reduced to partitioned and related groupings.

This is the rationale for a hybrid partition and k-nn scheme. Note that unlike stacking, our hybrid models are not independently determined, but interact very strongly with one another. However, it must be demonstrated that these methods are in fact complementary, preserving the strengths of the partitioning schemes while compensating for the weaknesses that would be introduced if constant values were used for each region. With respect to model combination, two principal questions need to be addressed by empirical experimentation:

- Are results improved relative to using each model alone?

- Are these methods competitive with alternative regression methods?

## 7. Results

Experiments were conducted to assess the competitiveness of rule-based regression compared to other procedures (including less interpretable ones), as well as to evaluate the performance of the integrated partition and k-nn regression method. Experiments were performed using seven datasets, six of which are described in previous studies (Quinlan, 1993). In addition to these six datasets, new experiments were done on a very large telecommunications application, which is labeled pole. In each of the seven datasets, there was one continuous real-valued response variable. Experimental results are reported in terms of the MAD, as measured using 10-fold cross-validation. For pole, 5,000 cases were used for training and 10,000 for independent testing. The features from the different datasets were a mixture of continuous and categorical features. For pole, all 48 features were continuous. Descriptions of the other datasets can be found in the literature (Quinlan, 1993).[4] Table 1 summarizes the key characteristics of the datasets used in this study.

| Dataset | Cases | Vars |
|---------|-------|------|
| price | 159 | 16 |
| servo | 167 | 19 |
| cpu | 209 | 6 |
| mpg | 392 | 13 |
| peptide | 431 | 128 |
| housing | 506 | 13 |
| pole | 15000 | 48 |

Table 1: Dataset Characteristics

Table 2 summarizes the original results reported (Quinlan, 1993). These include model-trees (MT), which are regression trees with linear fits at the terminal nodes; neural nets (NNET); 3-nearest neighbors (3-nn); and the combined results of model-trees and 3-nearest neighbors (MT/3-nn).[5]

Table 3 summarizes the additional results that we obtained. These include the CART regression tree (RT); 5-nearest neighbors with euclidean distance (5-nn); rule regression using Swap-1; rule regression with 5-nn applied to the rule region (Rule/5-nn); and MARS. 5-nn was used because the expectation is that the nearest neighbor method incrementally improves a constant-value region when the region has a moderately large sample of neighbors to average.

For the rule-based method, the parameter $m$, the number of pseudo-classes, must be determined. This can be found using cross-validation or independent test cases (in our experiments, cross-validation was used). Figure 6 represents a typical plot of the relative error vs. the number of pseudo-classes (Weiss & Indurkhya, 1993b). As the number of partitions increases, results improve until they reach a relative plateau and deteriorate somewhat. Similar complexity plots can be found for other models, for example neural nets (Weiss & Kapouleas, 1989).

The MARS procedure has several adjustable parameters.[6] For the parameter $mi$, values tried were 1 (additive modeling), 2, 3, 4 and number of inputs. For $df$, the default value of 3.0 was tried as well the optimal value estimated by cross-validation. The parameter $nk$ was varied from 20 to 100 in steps of 10. Lastly, both piece-wise linear as well as piece-wise cubic solutions were tried. For each of the above setting of the parameters, the cross-validated accuracy was monitored, and the value for the best MARS model is reported.

For each method, besides the MAD, the relative error is also reported. The relative error is simply the estimated true mean absolute distance (measured by cross-validation) normalized by the initial mean absolute distance from the median. Analogous to classifi-cation, where predictions must have fewer errors than simply predicting the largest class, in regression too we must do better than the average distance from the median to have meaningful results.

---

4. The peptide dataset is a slightly modified version of the one Quinlan refers to as *lhrh-att* in his paper. In the version used in our experiments, cases with missing values were removed.

5. Because peptide was a slightly modified version of the *lhrh-att* dataset, the result listed is one that was provided by Quinlan in a personal communication.

6. The particular program used was MARS 3.5.

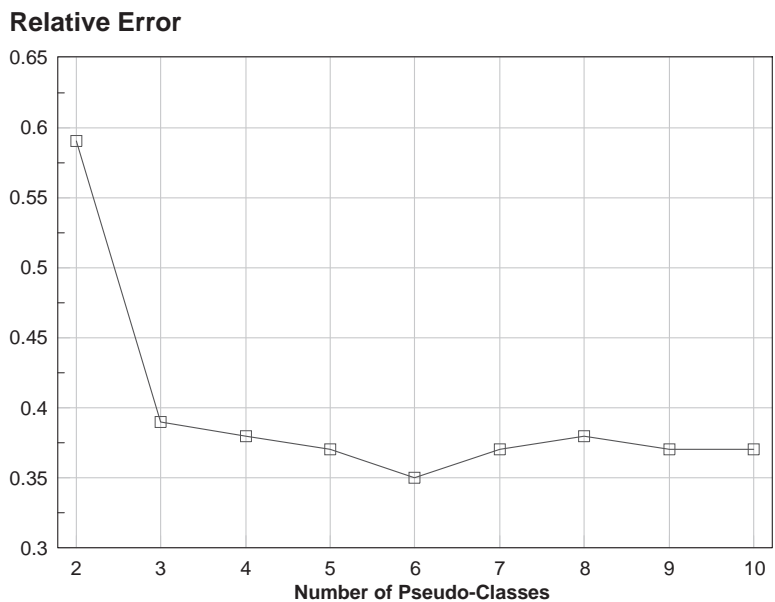

Figure 6: Prototypical Performance for Varying Pseudo-Classes

| Dataset | MT | NNET | 3-nn | MT/3-nn |
|---|---|---|---|---|
| price | 1562 | 1833 | 1689 | 1386 |
| servo | .45 | .30 | .52 | .30 |
| cpu | 28.9 | 28.7 | 34.0 | 28.1 |
| mpg | 2.11 | 2.02 | 2.72 | 2.18 |
| peptide | .95 | - | - | - |
| housing | 2.45 | 2.29 | 2.90 | 2.32 |

Table 2: Previous Results

In comparing the performance of two methods for a dataset, the standard error for each method was independently estimated, and the larger one was used in comparisons. If the difference in performance was greater than 2 standard errors, the difference was considered statistically significant. As with any significance test, one must also consider the overall pattern of performance and the relative advantages of competing solutions (Weiss & Indurkhya, 1994).

For each dataset, Figure 7 plots the relative best error found by the ratio of the best reported result to each model's result. A relative best error of 1 indicates that the result is the best reported result for any regression model. The model results that are compared to the best results are for regression rules, 5-nn, and the mixed model. The graph indicates trends across datasets and helps assess the overall pattern of performance. In this respect, both Rule and Rule/5nn exhibit excellent performance across many applications.

| Dataset | RT | | 5-nn | | Rule | | Rule/5-nn | | MARS | |
|---------|-----|------|-----|------|-----|------|-----|------|-----|------|
| | MAD | Error | MAD | Error | MAD | Error | MAD | Error | MAD | Error |
| price | 1660 | .40 | 1643 | .40 | 1335 | .32 | 1306 | .31 | 1559 | .38 |
| servo | .195 | .21 | .582 | .63 | .235 | .25 | .227 | .24 | .212 | .23 |
| cpu | 30.5 | .39 | 29.4 | .38 | 27.62 | .35 | 26.32 | .34 | 27.29 | .35 |
| mpg | 2.28 | .35 | 2.14 | .33 | 2.17 | .33 | 2.04 | .31 | 1.94 | .30 |
| peptide | .97 | .46 | .95 | .45 | .86 | .40 | .86 | .40 | .98 | .46 |
| housing | 2.74 | .42 | 2.77 | .42 | 2.51 | .38 | 2.35 | .36 | 2.24 | .34 |
| pole | 4.10 | .14 | 5.91 | .20 | 3.76 | .13 | 3.70 | .12 | 7.41 | .25 |

Table 3: Performance of Additional Methods

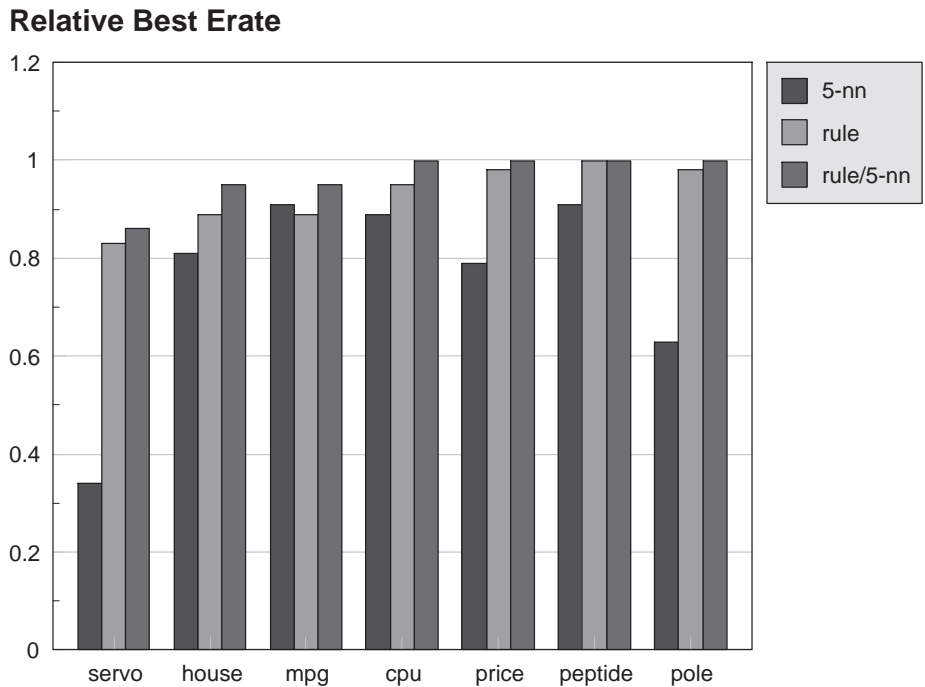

Figure 7: Relative Best Erates of 5-nn, Rules, and Rule/5-nn

These empirical results allow us to consider several relevant questions regarding rule-based regression:

1. *How does rule-based regression perform compared to tree-based regression?* Comparing the results for Rule with RT, one can see that except for servo, Rule does consistently better than RT on all the remaining six datasets. The difference in performance also tests as significant. The results of the significance tests, and the general trend (which can be seen visually in Figure 7) leads us to conclude that rule-based regression is definitely competitive to trees and often yields superior performance.

2. *Does integrating 5nn with rules lead to improved performance relative to using each model alone?* A comparison of Rule/5nn with 5nn shows that for all datasets, Rule/5nn is significantly better. In comparing Rule/5nn with Rule, the results indicate that for three datasets (mpg, pole and housing), Rule/5nn was significantly better than Rule, and for the remaining three datasets both were about the same. The overall pattern of performance also appears to favor Rule/5nn over Rule. Thus the empirical results indicate that our method improved results relative to using each model alone. The general trend can be seen in Figure 7.

3. *Are the new methods competitive with alternative regression methods?* Among the previous reported results, MT/3nn is the best performer. Other alternatives to consider are: Regression Trees (RT) and MARS. None of these three methods were significantly better than Rule/5nn on any of the datasets under consideration except for RT doing significantly better on servo. Furthermore, Rule/5nn was significantly better than MT/3nn on three of five datasets (servo, cpu and mpg) on which comparison is possible. The overall trend also is in favor of Rule/5nn. Comparing RT to Rule/5nn, we find that except for servo, Rule/5nn is significantly better than RT on all the remaining datasets. Comparing MARS to Rule/5nn, we find that for three of the datasets (price, peptide and pole), Rule/5nn is significantly better. Hence the empirical results overwhelmingly suggest that our new method is competitive with alternative regression methods, with hints of superiority over some methods.

## 8. Discussion

We have considered a new model for rule-based regression and provided comparisons with tree-based regression. For many applications, strong explanatory capabilities and high dimensional feature selection can make a DNF model quite advantageous. This is particularly true for knowledge-based applications, for example equipment repair or medical diagnosis, in contrast to pure pattern recognition applications such as speech recognition.

While rules are similar to trees, the rule representation is potentially more compact because the rules are not mutually exclusive. This potential of finding a more compact solution can be particularly important for problems where model interpretation is crucial. Note that the space of all rules includes the space of all trees. Thus, if a tree solution is the best, theoretically the rule induction procedure has the potential to find it.

In our experiments, the regression rules generally outperformed the regression trees. Fewer constant regions were required and the estimated error rates were generally lower. Finding the DNF regions was substantially more computationally expensive for the regression rules than the regression trees. For the regression rules, fairly complex optimization techniques were necessary. In addition, experiments must be performed to find the appropriate number of pseudo-classes. This is more a matter of scale: scale of the application versus the scale of available computing. Excluding the telecommunications application, none of the cited applications takes more than 15 minutes of cpu time on a SS-20 for a sin-gle pseudo-classification problem and a full cross-validation.[7] As computing power increases the timing distinction is less important. Even a small percentage gain can be quite valuable for the appropriate application (Apté, Damerau, & Weiss, 1994) and computational requirements are a secondary factor.

We have provided results on several real-world datasets. Mostly, these involve non-linear relationships. One may wonder how the rule-based method would perform on data with obvious linear relationships. In our earlier experiments with data exhibiting linear relationships (for example, the drug study data (Efron, 1988)), the rule-based solutions did slightly better than trees. However, the true test is real-world data which, often involve complex non-linear relationships. Comparisons with alternative models can help assess the effectiveness of the new techniques.

Looking at Figure 7 and Tables 2 and 3, we see that the pure rule-based solutions are competitive with other models. Additional gains are made when rules are used not for obtaining the function values directly, but instead used to find the relevant cases which are then used to compute the function value. The results of these experiments support the view that this strategy of combining different methods can improve predictive performance. Strategies similar to ours have been applied before for classification problems (Ting, 1994; Widmer, 1993) and similar conclusions were drawn from those results. Our results indicate that the strategy is useful in the regression context too. Our empirical results also support the contention that for regression, partitioning methods and nearest neighbor methods are complementary. A solution can be found by partitioning alone, and then the incremental improvement can be observed when substituting the average $y$ of the k-nearest neighbors for the median $y$ of a partition. From the perspective of nearest neighbor regression methods, the sample cases are compartmentalized, simplifying the table lookup for a new case.

While not conclusive, there are hints that our combination strategy is most effective for small to moderate samples: it is likely that when the sample size grows large, increased numbers of partitions, in terms of rules or terminal nodes, can compensate for having single constant-valued regions. This conjecture is supported by the large-sample pole application, where the incremental gain for the addition of k-nn is small.[8]

In our experiments we used k-nn with k=5. Depending on the application, a different value of k might produce better results. The optimal value might be estimated by cross-validation in a strategy that systematically varies k and picks the value that gives the best results overall. However, it is unclear whether the increased computational effort will result in any significant performance gain.

Another practical issue with large samples is the storage requirement: all the cases must be stored. This can be a serious drawback in real-world applications with limited memory. However, we tried experiments in which the cases associated with a partition are replaced by a fewer number of "typical cases". This results in considerable savings in terms of storage requirements. Results are slightly weaker (though not significantly different).

It would appear that further gains might be obtained by restricting the k-nn to consider only those features that appear in the path to the leaf node under examination. This might seem like a good idea because it attempts to ensure that only features that are relevant to the cases in the node, are used in the distance calculations. However, we found results for this to be weaker.

---

7. A 10-fold cross-validation requires solving a problem essentially 11 times: once on all training cases and 10 times for each group of test cases.

8. Although small, this difference tests as significant because the sample is large.

A number of regression techniques have been presented by others to demonstrate the advantages of combined models. Most of these combine methods that are independently invoked. Instead of a typical election where there is one winner, the alternative models are combined and weighted. These combination techniques have the advantage that the outputs of different models can be treated as independent variables. They can be combined in a form of post-processing, after all model outputs are available.

In no way do we contradict the value of these alternative combination techniques. Both approaches show improved results for various applications. We do conclude, however, that there are advantages for more complex regression procedures that dynamically mix the alternative models. These procedures may be particularly strong when there is a fundamental rationale for choice of methods such as partitioning methods, or when properties of the combined models must be preserved.

We have presented the regression problem with one output variable. This is the classical form for linear models and regression trees. The issue of multiple outputs has not been directly addressed although such extensions are feasible. This issue and further experimentation await future work. Our model of regression can provide a basis for these efforts, while leveraging current strong methods in classification rule induction.

# References

Apté, C., Damerau, F., & Weiss, S. (1994). Automated Learning of Decison Rules for Text Categorization. *ACM Transactions on Office Information Systems, 12*(3), 233–251.

Breiman, L. (1993). Stacked regression. Tech. rep., U. of CA. Berkeley.

Breiman, L., Friedman, J., Olshen, R., & Stone, C. (1984). *Classification and Regression Tress*. Wadsworth, Monterrey, Ca.

Clark, P., & Niblett, T. (1989). The CN2 induction algorithm. *Machine Learning, 3*, 261–283.

Craven, P., & Wahba, G. (1979). Smoothing noisy data with spline functions. estimating the correct degree of smoothing by the method of generalized cross-validation. *Numer. Math., 31*, 317–403.

Efron, B. (1988). Computer-intensive methods in statistical regression. *SIAM Review, 30*(3), 421–449.

Fayyad, U., & Irani, K. (1992). The attribute selection problem in decision tree generation. In *Proceedings of AAAI-92*, pp. 104–110 San Jose.

Friedman, J. (1991). Multivariate adaptive regression splines. *Annals of Statistics, 19*(1), 1–141.

Friedman, J., & Stuetzle, W. (1981). Projection pursuit regression. *J. Amer. Stat. Assoc., 76*, 817–823.

Girosi, F., & Poggio, T. (1990). Networks and the best approximation property. *Biological Cybernetics, 63*, 169–176.

Hartigan, J., & Wong, M. (1979). A k-means clustering algorithm, ALGORITHM AS 136. *Applied Statistics, 28*(1).

Hastie, T., & Tibshirani, R. (1990). *Generalized Additive Models*. Chapman and Hall.

Jacoby, S., Kowalik, J., & Pizzo, J. (1972). *Iterative Methods for Non-linear Optimization Problems*. Prentice-Hall, New Jersey.

Kirpatrick, S., Gelatt, C., & Vecchi, M. (1983). Optimization by simulated annealing. *Science, 220*, 671.

LeBlanc, M., & Tibshirani, R. (1993). Combining estimates in regression and classification. Tech. rep., Department of Statistics, U. of Toronto.

Lebowitz, M. (1985). Categorizing numeric information for generalization. *Cognitive Science, 9*, 285–308.

Lin, S., & Kernighan, B. (1973). An efficient heuristic for the traveling salesman problem. *Operations Research, 21*(2), 498–516.

McClelland, J., & Rumelhart, D. (1988). *Explorations in Parallel Distributed Processing*. MIT Press, Cambridge, Ma.

Michalski, R., Mozetic, I., Hong, J., & Lavrac, N. (1986). The multi-purpose incremental learning system AQ15 and its testing application to three medical domains. In *Proceedings of AAAI-86*, pp. 1041–1045 Philadelphia, Pa.

Quinlan, J. (1986). Induction of decision trees. *Machine Learning, 1*, 81–106.

Quinlan, J. (1987). Simplifying decision trees. *International Journal of Man-Machine Studies, 27*, 221–234.

Quinlan, J. (1993). Combining instance-based and model-based learning. In *International Conference on Machine Learning*, pp. 236–243.

Ripley, B. (1993). Statistical aspects of neural networks. In *Proceedings of Seminair Europeen de Statistique* London. Chapman and Hall.

Scheffe, H. (1959). *The Analysis of Variance*. Wiley, New York.

Ting, K. (1994). The problem of small disjuncts: Its remedy in decision trees. In *Proceedings of the 10th Canadian Conference on Artificial Intelligence*, pp. 91–97.

Weiss, S., & Indurkhya, N. (1993a). Optimized Rule Induction. *IEEE Expert, 8*(6), 61–69.

Weiss, S., & Indurkhya, N. (1993b). Rule-based regression. In *Proceedings of the 13th International Joint Conference on Artificial Intelligence*, pp. 1072–1078.

Weiss, S., & Indurkhya, N. (1994). Decision tree pruning: Biased or optimal?. In *Proceedings of AAAI-94*, pp. 626–632.

Weiss, S., & Kapouleas, I. (1989). An empirical comparison of pattern recognition, neural nets, and machine learning classification methods. In *International Joint Conference on Artificial Intelligence*, pp. 781–787 Detroit, Michigan.

Weiss, S., & Kulikowski, C. (1991). *Computer Systems that Learn: Classification and Prediction Methods from Statistics, Neural Nets, Machine Learning, and Expert Systems*. Morgan Kaufmann.

Widmer, G. (1993). Combining knowledge-based and instance-based learning to exploit qualitative knowledge. *Informatica*, *17*, 371–385.

Wolpert, D. (1992). Stacked generalization. *Neural Networks*, *5*, 241–259.